\documentclass[runningheads]{llncs}

\usepackage{eccv}

\usepackage{eccvabbrv}

\usepackage{graphicx}
\usepackage{booktabs}

\usepackage[accsupp]{axessibility}  % Improves PDF readability for those with disabilities.

\usepackage{hyperref}

\usepackage{orcidlink}

\usepackage[ruled,vlined,linesnumbered]{algorithm2e}
\usepackage{subcaption}
\usepackage{adjustbox}
\usepackage[table]{xcolor}
\usepackage{multirow}
\usepackage{verbatim}
\usepackage{float}

\begin{document}

\title{DA-MergeLoRA: Hypernetwork-Based LoRA Merging for Few-Shot Test-Time Domain Adaptation} 

\titlerunning{DA-MergeLoRA: Hypernetwork-Based LoRA Merging for FSTT-DA}

\author{
Siobhan Reid\inst{1}\orcidlink{0009-0000-2278-609X}\and
Zhixiang Chi\inst{2}\orcidlink{0000-0003-4560-4986}\and
Li Gu \inst{1,3}\and
Omid Reza Heidari \inst{1}\orcidlink{0009-0000-3518-643X}\and
Ziqiang Wang \inst{1,3}\orcidlink{0000-0002-4083-5411}\and
Yang Wang \inst{1,3}\orcidlink{0000-0001-9447-1791} 
}

\authorrunning{S.~Reid et al.}

\institute{Concordia University, Montreal QC H3G 1M8, Canada \and
University of Toronto, Toronto ON M5S 1A1, Canada
\and 
Mila – Quebec AI Institute, Montreal QC H2S 3H1, Canada}

\maketitle

\begin{abstract}
Few-shot Test-Time Domain Adaptation (FSTT-DA) seeks to adapt models to novel domains using only a handful of unlabeled target samples. This setting is more realistic than typical domain adaptation setups, which assume access to target data during source training. However, prior FSTT-DA approaches fail to effectively leverage source domain-specific knowledge, relying on shallow batch normalization updates, prompt-based methods that treat the model as a black box, or ensembling strategies that do not capture cross-domain relationships. To address these limitations, we introduce a new FSTT-DA framework that integrates LoRA fine-tuning with model merging. In our approach, separate LoRA modules are fine-tuned on CLIP’s vision encoder for each source domain. Since LoRA modifies only a small fraction of the model’s parameters, it retains the base model’s generalized knowledge while internally learning domain-specific features. To adapt the learned knowledge to a specific target domain, we propose a hypernetwork trained via meta-learning that generates per-column merging factors to combine LoRA modules. Given a small batch of target images, the hypernetwork produces merging weights that fuse source LoRA modules into a single adapted representation. Our results demonstrate state-of-the-art performance across various domain adaptation datasets. Our code is publicly available at \url{https://github.com/nahbois4321/DA-MergeLoRA}.

\end{abstract}

\section{Introduction}

Machine learning models typically assume that training and test data are identically distributed; when the test distribution differs (domain shift), performance degrades \cite{dg1,dg2}. Test-Time Adaptation (TTA) addresses domain shift by adapting a model at inference time, typically operating in either an offline or online mode~\cite{chi2025plug,wu2026talon,wang2024distribution,jiang2026pointmac}. In the offline setting, the model is given a large corpus of unlabeled target data and performs multi-step unsupervised adaptation before inference. In the online setting, the model receives a stream of target samples and updates itself incrementally, often on a per-sample or per-mini-batch basis, throughout deployment \cite{tent}. Both modes assume access to many target samples and allow repeated adaptation steps.

In contrast, Few-Shot Test-Time Domain Adaptation (FSTT-DA) assumes that the model receives only a single small batch of unlabeled target images at deployment and must perform a one-shot adaptation step before processing the entire target domain, with no further updates \cite{wu2024mabn}. In this setting, the model can be trained on labeled data from multiple source domains during development, but the target domain remains completely unseen until test time and is available only as a small unlabeled batch for adaptation. This differs from typical multi-source domain adaptation \cite{domainnet}, which assumes access to the unlabeled target domain during source training for alignment or distribution matching. 

FSTT-DA arises in real-world applications where data are limited, inference-time overhead must be minimal, or adaptation using target data is restricted by privacy or safety constraints. For example, medical imaging models deployed in new hospitals face domain shifts; however, regulations can prevent iterative or large-scale adaptation using patient data. Likewise, robots entering novel environments typically receive only a handful of initial observations before they must begin downstream tasks with minimal inference-time overhead. FSTT-DA is therefore a practical yet more challenging setting than typical TTA.

This setting poses two key difficulties: \textbf{(CI)} extracting domain-specific cues from diverse source domains without harming cross-domain generalization, and \textbf{(CII)} transferring that knowledge to a new target domain under severe data and label scarcity. Despite recent progress, current FSTT-DA methods remain limited. BatchNorm-based adaptation, MABN \cite{wu2024mabn}, is efficient but restricted to re-estimating BatchNorm statistics and affine parameters; with small test batches, these estimates become noisy and lead to unstable transfer. Meta-DMoE \cite{metadmoe} employs ensemble/distillation strategies to train multiple source-domain experts via a teacher–student pipeline, incurring substantial compute, scaling poorly with the number of sources, and limiting cross-domain sharing. Prompt-based approaches \cite{vdpg,l2c} treat the backbone model as a black box and adjust only external prompts, leaving internal representations fixed and restricting knowledge transfer. As a result, existing methods make shallow updates, scale poorly, or restrict knowledge sharing, and ultimately fall short of the two key requirements \textbf{(CI–CII)}.

To overcome \textbf{(CI)}, we draw inspiration from recent work showing that inserting lightweight low-rank adapters (LoRA)~\cite{lora} in parallel with model weights can effectively capture domain-specific information~\cite{liuvida}. We attach separate sets of LoRAs to a frozen CLIP backbone to model knowledge from multiple source domains. This design provides more expressive adaptation than shallow BatchNorm updates \cite{wu2024mabn} and allows internal knowledge steering, unlike black-box prompting methods such as VDPG \cite{vdpg} and L2C \cite{l2c}. Moreover, unlike ensemble methods that require a full model per source domain, our approach enables cross-domain knowledge sharing while retaining the generalization of the base model.

To address \textbf{(CII)}, we adopt parameter-space model merging to integrate domain-specific LoRA modules into a single model adapted to the target domain. Unlike prompt-based methods \cite{vdpg,l2c}, merging updates internal weights and aligns both low- and high-level features. Compared to ensembling \cite{metadmoe}, it avoids running multiple experts and produces one model with fixed memory and a single forward pass. Beyond efficiency, merging promotes direct knowledge sharing across domains and avoids the instability of retraining or distillation in the few-shot setting.

To further motivate our approach, parameter-space model merging has been successfully applied in several other areas. One line of work merges task-specific models into a single multi-task generalist model \cite{ties, regmean, fishermerge, isoc, knots, tsvm}. Unlike these approaches, which aim to unify different tasks into one generalist model, our goal is to create a specialized domain-specific model tailored to a novel visual domain. Another line of work merges pairs of target LoRAs at test time to produce mixed style-content LoRAs in diffusion models \cite{lorarar, ziplora}. In contrast, our setting has no target LoRA available; instead, we must synthesize a domain-specific LoRA by combining source-domain LoRAs guided by only a few unlabeled target samples. Model merging has also been explored for NLP task adaptation \cite{mole, lorahub, llmmerge}, where each LoRA corresponds to a different supervised task (QA, translation, NLI, etc.). Unlike previous task-centric merging, our work focuses on a fixed-task classification setting that undergoes visual distribution shifts across domains, motivating a merging strategy tailored for domain adaptation rather than task-specific tuning.

To this end, we introduce \textbf{DA-MergeLoRA}, an FSTT-DA framework that reformulates adaptation as parameter-space merging. For each source domain, we fine-tune a separate LoRA module on the CLIP vision encoder \cite{clip}, while keeping the base parameters frozen to preserve cross-domain generalization. At test time, given a few unlabeled target images, a hypernetwork predicts merging weights to combine the source-domain LoRAs into a single target LoRA. The hypernetwork is conditioned on both target images and the pool of source LoRA weights, and is trained in a meta-learning fashion to learn effective merging policies. Empirically, our approach achieves state-of-the-art results across multiple FSTT-DA benchmarks, including a \textbf{+1.24\%} gain in average accuracy on DomainNet, \textbf{+2.70\%} on Camelyon17, and \textbf{+11.40\%} in worst-case accuracy on FMoW.

Our work makes the following contributions: (i) Novel FSTT-DA framework: we formulate FSTT-DA as a parameter-space merging problem, yielding a single adapted model. (ii) Hypernetwork architecture: We introduce a meta-learned lightweight hypernetwork that generates per-column merge weights to combine multiple source LoRA modules into a specialized, domain-specific target model, conditioned on a small batch of unlabeled target data, enabling effective FSTT-DA. (iii) State-of-the-art results: Our approach achieves SOTA on the DomainNet and WILDS benchmarks, empirically demonstrating that parameter-space merging is an effective approach for FSTT-DA. Together, these contributions move beyond prior FSTT-DA methods (batch normalization, prompt-based, and distillation ensembles) toward modular internal source-domain representations and an adaptive parameter-space mechanism for effective test-time adaptation.

\section{Related Work}

\subsubsection{Test-time domain adaptation.} Distribution shift between training and test data often degrades performance when models are deployed to new domains \cite{dg3}. Domain adaptation algorithms address this by learning features shared between source and target domains, enabling better generalization \cite{mmd,dann,dsn}. Test-time adaptation (TTA) considers the setting where no target-domain data is available during model development \cite{tent}. After deployment to a new domain, the model may adapt in an offline manner, where the model is trained on all unlabeled target data before inference, or in an online manner, where the model is continually adapted to streaming test data \cite{tta}. Common TTA strategies include entropy minimization \cite{tent, wang2024distribution}, pseudo-labels \cite{pseudoda}, auxiliary tasks \cite{auxtasks,liu2022towards}, and contrastive learning \cite{conda}.

\subsubsection{Few-shot test-time domain adaptation.} In FSTT-DA, models adapt to unseen domains using only a batch of unlabeled samples at inference \cite{metadmoe}. Prior works approach this challenge in different ways. Meta-DMoE distills knowledge from a mixture of source-domain experts to a new target model via a meta-learned Transformer aggregator \cite{metadmoe}. MABN adapts batch-normalization affine parameters using a self-supervised auxiliary branch \cite{wu2024mabn}. VDPG generates domain-specific visual prompts from a knowledge bank conditioned on target samples \cite{vdpg}. L2C enhances VDPG by attaching a parallel branch to CLIP, learning directly from dataset-specific input knowledge and text semantics \cite{l2c}.

\subsubsection{Foundation models.} Recent FSTT-DA methods,VDPG \cite{vdpg} and L2C \cite{l2c}, use CLIP as a frozen backbone. CLIP is a vision–language model whose image and text encoders are trained using contrastive learning so that paired images and captions have high cosine similarity. Classification then selects the prompt “a photo of a \textless classname\textgreater” with the highest similarity \cite{clip}. Vision–language models are appealing for domain adaptation because of their strong out-of-distribution generalization. However, prior research shows that full fine-tuning of CLIP can degrade its generalization capabilities \cite{ftclip}, motivating parameter-efficient methods such as prompt tuning \cite{vdpg} and LoRA \cite{lora}, which keep the backbone frozen while learning domain-specific features. Similar to VDPG and L2C, our work builds on CLIP, but differs by encoding domain knowledge directly in LoRA modules, enabling richer and more flexible adaptation than prompt- or batch normalization-based approaches.

\subsubsection{Model merging.} Model-merging methods aim to combine multiple task-specific models into a single general-purpose model while mitigating task interference. Examples include parameter averaging \cite{avg1,avg2}, Task Arithmetic \cite{taskarithmetic}, Fisher Merging \cite{fishermerge}, RegMean \cite{regmean}, and structured merging methods such as TIES \cite{ties}, which resolves sign conflicts and averages only aligned updates. Several recent advances further improve alignment and reduce interference: KnOTS \cite{knots} applies singular value decomposition (SVD) to concatenated LoRA updates before merging task-aligned components; Iso-C / Iso-CTS \cite{isoc} use SVD and isotropic scaling to identify shared and task-specific directions; and TSV-M \cite{tsvm} extracts low-rank task directions via SVD and orthogonalizes them across tasks. These approaches all aim to produce a single multi-task model that preserves performance across the original training tasks. In contrast, our method focuses on generating a specialized domain-specific model, as opposed to a generalist model. Merging in prior work is unconditional (merge rules do not depend on target data) whereas our merging is conditional and meta-learned, driven by a target-domain embedding extracted from a few unlabeled target images.

\subsubsection{LoRA merging.} LoRA fine-tuning inserts low-rank adapters into frozen foundation models, enabling efficient adaptation with fewer trainable parameters. LoRA merging has recently been applied across different domains \cite{lorahub,lorarar,sdlora,icmfusion,regcl,elrea,moa,text2lora}. In diffusion models, LoRA.rar \cite{lorarar} and ZipLoRA \cite{ziplora} merge pairs of content and style LoRAs for personalized image generation, while MoLE \cite{mole} trains a gating network to combine multiple content-specific LoRAs into a mixed-content adapter. These approaches all assume that target LoRA modules already exist and focus on fusing them coherently. In contrast, we assume no target LoRA is available; our method must construct a target LoRA by merging source-domain LoRAs, guided only by a few unlabeled target images. Several recent works explore LoRA merging and LoRA generation in NLP and vision. LoRAHub \cite{lorahub} learns to merge LoRA modules for NLP task transfer but requires few-shot labeled examples, whereas our adaptation uses only unlabeled target data. SD-LoRA \cite{sdlora} incrementally incorporates new LoRA modules for continual learning by decoupling their direction and magnitude. Text-to-LoRA \cite{text2lora} uses a hypernetwork to generate LoRA weights from natural-language task descriptions. ICM-Fusion \cite{icmfusion} fuses LoRA modules using a VAE-based latent representation, and RegCL \cite{regcl} merges LoRA adapters for continual-learning domain-specific image segmentation tasks. These diverse successes from other research areas further motivate exploring LoRA merging within the FSTT-DA setting.

\subsubsection{Meta-learning.} Meta-learning is a training paradigm that enables a model to quickly adapt to new tasks or domains \cite{maml}. Previous FSTT-DA methods, such as VDPG \cite{vdpg}, employ meta-learning to generate domain-specific visual prompts from a knowledge bank. In contrast to these prompt-based approaches which leave internal representations fixed, we use meta-learning to train a hypernetwork that predicts per-column merging weights. This allows our method to dynamically synthesize a specialized target LoRA module in the parameter space.

\section{Method}

\begin{figure}[t]  
    \centering
    \includegraphics[clip, trim=0.3cm 0.1cm 0.3cm 0.1cm, width=1.0\linewidth]
    {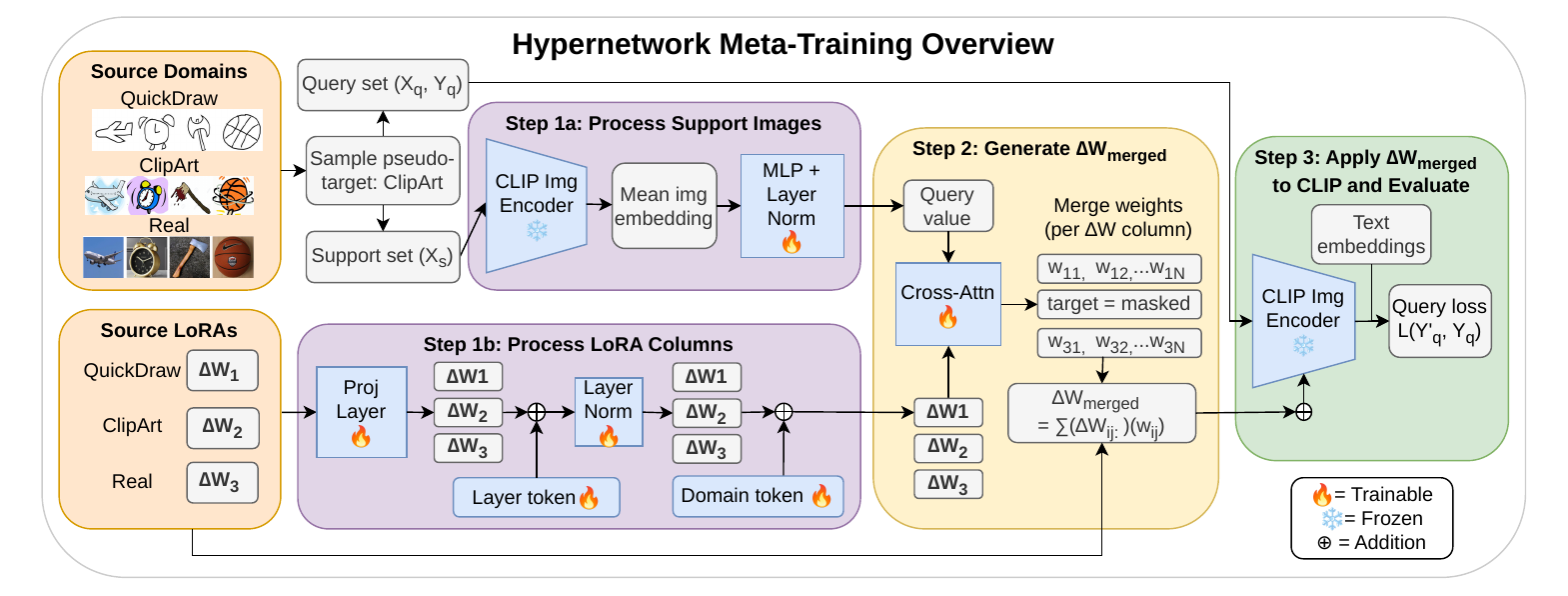} %
    \caption{
\textbf{Overview of our hypernetwork training procedure.} Given pre-trained source LoRA modules, we meta-train a hypernetwork to dynamically predict optimal weights to merge the source LoRA modules into a target-domain specific model. At each meta-training step, one source domain is randomly sampled as the pseudo-target, while the remaining domains serve as source data. Cross-attention is computed between the processed pseudo-target support images and source LoRA columns to generate column-wise merge weights, yielding $\Delta W_{\text{merged}}$. Finally, the merged model is applied to the frozen CLIP backbone and evaluated on query images to compute the meta-training loss.
    }
    \label{fig:hypernet_overview}
\end{figure}

\subsubsection{Problem setting.} 
In this work, we follow the same problem setting as previous FSTT-DA methods \cite{metadmoe, vdpg, l2c}. The training set consists of $N$ labeled source domains $\mathcal{D}_s = \{\mathcal{D}_s^n\}_{n=1}^N$, where each source domain $\mathcal{D}_s^n = (x_s, y_s)^n$ contains image--label pairs $(x,y)$. 
The test set includes $M$ target domains $\mathcal{D}_t = \{\mathcal{D}_t^m\}_{m=1}^M$, 
where each target domain $\mathcal{D}_t^m = (x_t)^m$ provides only unlabeled images $x_t$. 
We assume a distribution shift exists between any source and target domain, while all domains share the same label space. FSTT-DA trains exclusively on the source domains, without access to target data. 
At test time, when an unseen target domain $\mathcal{D}_t^m$ is encountered (deployed in a new environment), the model must adapt to this domain using only a few-shot batch of unlabeled samples. The adapted model is then evaluated on the entire target domain.

\subsubsection{Overview.} In this work, we develop a hypernetwork that generates merging weights to combine LoRA modules, conditioned on a batch of unlabeled target images. Sec.~\ref{sec:lora_training} outlines the training of individual LoRA modules. Sec.~\ref{sec:lora_hypernetwork} presents the architecture of the LoRA hypernetwork, which produces per-column merging weights. Sec.~\ref{sec:hypernetwork_training} details the meta-learning setup used to train the hypernetwork and the inference procedure.

\subsection{LoRA Module Training}
\label{sec:lora_training}
To capture domain-specific knowledge, a LoRA model is trained on each source domain using CLIP \cite{clip}. LoRA adapts the vision encoder by adding trainable low-rank matrices to its attention and projection weight matrices \cite{lora}. Instead of fine-tuning the weight matrix $W \in \mathbb{R}^{d \times k}$, LoRA replaces the weight update with a low-rank approximation, learned through two smaller matrices $A$ and $B$:
\begin{equation}
W' = W + \Delta W, \quad \Delta W = BA
\end{equation}
where $A \in \mathbb{R}^{r \times k}$ and $B \in \mathbb{R}^{d \times r}$ with $r \ll \min(d,k)$. The rank $r$ controls the trade-off between capacity and efficiency. During training, only $A$ and $B$ are updated, while $W$ remains frozen. For an input $h$, the forward pass becomes:
\begin{equation}
h' = W h + BA h = W h + \Delta W h
\end{equation}

The backbone CLIP model is frozen throughout training, and only the LoRA parameters are updated. Before training, class-specific text prompt embeddings (e.g., “a photo of a \textless classname\textgreater”) are precomputed using CLIP’s text encoder. During training, the model uses the cross-entropy loss to learn to select the prompt embedding with the highest cosine similarity to the LoRA-modified image embedding.

\subsection{LoRA Merging Hypernetwork}
\label{sec:lora_hypernetwork}

A hypernetwork is a neural network that generates the weights or parameters of another network \cite{lorarar}. Here, we use a hypernetwork that learns to generate merging weights to combine the columns of LoRA matrices $\Delta W$ trained on different source domains. Prior work \cite{taskarithmetic} shows that models fine-tuned from a shared backbone (e.g., CLIP) remain compatible for arithmetic merging. The hypernetwork accepts as input a batch of target images and the LoRA $\Delta W$ columns from the different source domains, utilizing cross-attention to find the source columns that correlate best with the target images. Following LoRA.rar \cite{lorarar}, we employ per-column $\Delta W$ merging rather than per-model or per-layer merging. Theoretically, this grants a more flexible weighting mechanism to selectively emphasize or suppress feature subspaces relevant to the target domain.

\subsubsection{Offline LoRA preparation.} Before hypernetwork training begins, the full product matrices $\Delta W = BA$ for all source domains are pre-calculated offline for each transformer attention and projection layer. We explicitly merge the columns of the full $\Delta W$ product, as opposed to merging the separate low-rank matrices $A$ or $B$, because it allows for fine-grained merging and enables us to combine different LoRA models regardless of their rank $r$. For this column-wise merging, the hypernetwork generates weight vectors matching each layer's dimensionality ($d=768$ for CLIP ViT-B/16).

\subsubsection{Target image processing.} During each forward pass of the hypernetwork, the current batch of target domain images is processed before being used for cross-attention. The images, $X_{d}$, are passed through the frozen CLIP backbone to generate embeddings, and the mean embedding, $e_d$, is found:

\begin{equation}
e_{d} = \mathrm{mean}(\mathrm{EncodeImage}(X_{d}))
\end{equation}

The mean embedding is then passed through a small multi-layer perceptron ($\mathrm{MLP}$) network to reduce its dimension to $\mathbb{R}^{128}$ and extract domain-specific information. The MLP consists of two linear layers with a GELU activation and hidden dimension size of $\mathbb{R}^{256}$. The output from the MLP is then passed through a LayerNorm ($\mathrm{LN}$) to normalize the output:
\begin{equation}
e'_{d} = \mathrm{LN}(\mathrm{MLP}(e_d))
\end{equation}

\subsubsection{LoRA column processing.} Before LoRA $\Delta W$ columns are used for cross-attention to calculate merge weights, they are also processed. Firstly, they are passed through a projection layer, which maps their input dimension to a uniform dimension of $\mathbb{R}^{128}$. Additionally, we add three content tokens to the columns to help the hypernetwork learn more fine-grained and context-aware merging weights: 
(i) Layer index token: a learnable embedding $u_{\ell} \in \mathbb{R}^{128}$ representing the transformer layer $\ell$ the column came from; (ii) Sub-layer type token: a learnable embedding $v_{t} \in \mathbb{R}^{128}$ indicating whether the column comes from an attention (\texttt{qkv}) or projection (\texttt{proj}) site of the transformer layer. These two tokens act as positional embeddings; (iii) Domain token: for each source domain $d$, a learnable embedding $q_{d}$ is projected to $\mathbb{R}^{128}$ and added after normalization, providing a lightweight bias indicating which domain the column came from:
\begin{equation}
c^{\star} = \mathrm{LN}(\tilde{c} + u_{\ell} + v_{t}) + \alpha P_{\text{dom}} q_{d}
\end{equation}
Here, $\tilde{c} \in \mathbb{R}^{128}$ is the projected LoRA column, $P_{\text{dom}} \in \mathbb{R}^{128 \times E'}$ projects the domain embedding into the 128-dimensional space, and $\alpha$ is a small scaling factor that controls the influence of the domain token. The resulting $c^{\star} \in \mathbb{R}^{128}$ is the final context-enriched column representation used in cross-attention.

\subsubsection{Cross-attention.} Cross-attention is then used to produce the merge weights, where the query is the embedded target domain representation and the keys are the processed LoRA columns:
\begin{equation}
w^{(k)}_{\ell,t,c} = \mathrm{softmax}_{k}\!\left(\frac{Q \, K_{\ell,t,k,c}^{\top}}{\tau}\right)
\end{equation}
Here, $Q$ is the projected target-domain embedding, $K_{\ell,t,k,c}$ is the key vector for the LoRA column at transformer layer $\ell$, site $t$, domain $k$, and column index $c$, and $\tau$ is a temperature scaling factor. The softmax is taken along the domain axis $k$, producing normalized per-domain weights for each column. 

After the merge weights are generated using the hypernetwork, the LoRA matrices $\Delta W$ from the different source domains are combined together:
\begin{equation}
\big(\Delta W^{\mathrm{merge}}_{\ell,t}\big)_{:,c}
= \sum_{k=1}^{K} w^{(k)}_{\ell,t,c}\, \big(\Delta W^{(k)}_{\ell,t}\big)_{:,c}
\qquad c=1,\dots, C_{\ell,t}
\end{equation}
Here, $\Delta W^{(k)}_{\ell,t}$ denotes the LoRA update at transformer layer $\ell$ and site $t \in \{\text{attn}, \text{proj}\}$ for source domain $k$. $w^{(k)}_{\ell,t,c}$ is the merging weight for column $c$. $C_{\ell,t}$ is the number of columns at that $(\ell,t)$ site. $K$ is the number of source domains. After each merged $\Delta W$ is computed, they are applied to the frozen CLIP model to generate the new merged model.

\begin{algorithm}[t]
\tiny
\DontPrintSemicolon
\SetAlgoLined
\LinesNotNumbered
\SetAlgoNlRelativeSize{-1}

\KwIn{$\mathcal{D}_s$: source domains; $\{\Delta W^{(k)}\}_{k=1}^{K}$: frozen LoRA weight matrices from the K source domains;
frozen base weights $W$ (CLIP); trainable hypernetwork $\mathcal{H}$; step size $\alpha$}
\KwOut{Trained hypernetwork $\mathcal{H}$}

Randomly initialize $\theta$ (params of $\mathcal{H}$)\;
\While{not converged}{
  Sample pseudo-target domain index $d \sim \{1,\dots,K\}$\;
  Sample support/query from domain $d$: $(X_{S_d},Y_{S_d}), (X_{Q_d},Y_{Q_d})$\;
  Compute support embedding with frozen CLIP:
  $e_{S_d} \leftarrow \mathrm{mean}(\mathrm{EncodeImage}(X_{S_d}))$\;
  Set pseudo-target LoRA columns to zero: $\Delta W^{(d)} \leftarrow 0$\;
  Generate per-column merge weights using the hypernet: $w \leftarrow \mathcal{H}\!\left(e_{S_d}, \{\Delta W^{(k)}\}_{k=1}^{K}\right)$\;
  \ForEach{layer, $\ell$, layer type $t$, and column $c$}{
    Perform merge: $\big(\Delta W^{\mathrm{merge}}_{\ell,t}\big)_{:,c}
= \sum_{k=1}^{K} w^{(k)}_{\ell,t,c}\, \big(\Delta W^{(k)}_{\ell,t}\big)_{:,c}$\;
  }
  Apply merged LoRA matrices to base CLIP model: $W' \leftarrow W + \Delta W^{(\mathrm{merge})}$\;
  Evaluate on query set with merged model: $\hat{Y}_{Q_d} \leftarrow f_{W'}(X_{Q_d})$\;
  Calculate loss: $\mathcal{L} \leftarrow \mathrm{CE}(\hat{Y}_{Q_d}, Y_{Q_d})$\;
  Update hypernetwork parameters: $\theta_{\mathcal{H}} \leftarrow \theta_{\mathcal{H}} - \alpha \nabla_{\theta_{\mathcal{H}}}\mathcal{L}$\;
}
\caption{Overview of our hypernetwork meta-training algorithm.}
\label{alg:algo}
\end{algorithm}

\subsection{Hypernetwork Meta-Training and Inference}
\label{sec:hypernetwork_training}

\subsubsection{Training.} The hypernetwork is trained using a meta-learning framework, similar to~\cite{vdpg}. At each meta-iteration, one source domain is randomly selected as a pseudo-target domain to simulate test-time evaluation on novel domains. The LoRA columns of the pseudo-target domain are masked with zeros, and its merge weight is fixed to zero. From this pseudo-target domain, a support batch and a query batch are sampled. The support set images and the LoRA columns from the other source domains are passed to the hypernetwork, which generates the column-wise merge weights. The resulting merged LoRA matrices are applied to the frozen CLIP backbone, and the merged model is evaluated on the query set. A standard cross-entropy classification loss is computed on the query set predictions, and the resulting gradients are backpropagated end-to-end, from the merged model to the hypernetwork. The hypernetwork's parameters are trainable, while all other parameters are frozen (CLIP backbone and source LoRA models). The training procedure is detailed in Algorithm~\ref{alg:algo} and visualized in Fig.~\ref{fig:hypernet_overview}.

\subsubsection{Inference:} At test time, an unlabeled batch of images from the target domain and all source LoRA modules are passed to the hypernetwork. The target domain is novel and excluded from the hypernetwork training process. The hypernetwork outputs the merging weights, which are used to combine the source LoRA modules into a new model optimized for the target domain. After this initial one-step adaptation, the merged model is then evaluated on the entire target domain with no further updates.

\section{Experiments}

\subsubsection{Datasets.} We follow prior FSTT-DA work~\cite{vdpg,l2c} and evaluate on the WILDS and DomainNet benchmarks. We evaluate on three WILDS~\cite{wilds} benchmark classification datasets (iWildCam~\cite{iwildcam}, FMoW~\cite{FMOW}, and Camelyon17~\cite{camelyon}), as well as the WILDS PovertyMap~\cite{poverty} benchmark regression dataset. These datasets consist of real-world data with distribution shifts. We use the official WILDS evaluation metrics for each dataset (average accuracy, F1-score, worst-case (WC) accuracy, and WC Pearson correlation (r)) and the official train and out-of-distribution (OOD) test splits. We also evaluate on DomainNet~\cite{domainnet}, which contains roughly 600k images from 345 classes across six domains. We use the standard leave-one-out protocol for DomainNet, selecting one domain for testing and the remainder as source domains. Images are preprocessed using CLIP's standard train and test transforms. The PovertyMap~\cite{poverty} dataset uses 8-channel input images. We follow prior FSTT-DA work~\cite{vdpg,l2c} and use the first 3 out of 8 channels, so that its input is compatible with CLIP. Additionally, we use the English version of the class names for iWildCam.

\subsubsection{Baselines:} We compare our results against common CNN-based \cite{coral, irm, arm, metadmoe, wu2024mabn, mixup, mtl, segnet} and CLIP-based \cite{vdpg, l2c, doprompt, miro} domain adaptation methods, which include recent FSTT-DA algorithms (MABN \cite{wu2024mabn}, Meta-DMOE \cite{metadmoe}, VDPG \cite{vdpg}, and L2C \cite{l2c}). Additionally, we compare against zero-shot CLIP with no fine-tuning.  

\subsubsection{Training Details.} We use OpenAI's pretrained CLIP ViT-B/16 backbone~\cite{clip}. All LoRA models use a rank of 16 and $\alpha=32$. Learning rates are set between $3\times 10^{-4}$ to $5\times 10^{-4}$ for LoRA fine-tuning, and between $1\times 10^{-5}$ to $5\times 10^{-5}$ for hypernetwork training, with a batch size of 16 for both query and support sets. Hypernetwork models are trained for 1 to 5 epochs, except for PovertyMap, which uses an additional MLP regression head and is trained for 50 epochs. The text prompts used for training are dataset-specific, following the templates from L2C~\cite{l2c}; the mean embedding from these text prompts is used as the final text embedding. Additionally, following Meta-DMoE \cite{metadmoe}, the iWildCam source domains are grouped into 10 clusters for training. Detailed hyperparameter configurations and training procedures are provided in Appendix A.3 of the supplementary material.

\begin{table}[t]
\centering
\caption{Evaluation of our method (DA-MergeLoRA) on the (a) DomainNet dataset and (b) WILDS datasets. The best CNN and best ViT methods are bolded separately. Results are reported as mean (standard deviation) over three trials with different random seeds. The $^\dagger$ symbol denotes the official dataset metric. Our method achieves SOTA results on five of the six DomainNet domains and on three of the four official WILDS metrics.}
\captionsetup[subtable]{position=top}
% ---------- (b) DomainNet ----------
\begin{subtable}{\textwidth}
\centering
\footnotesize
\setlength{\tabcolsep}{2pt}
\renewcommand{\arraystretch}{1.08}
\begin{adjustbox}{width=\textwidth}
\begin{tabular}{@{}%
  >{\raggedright\arraybackslash}p{3.25cm}
  @{\hspace{4pt}}c
  *{7}{@{\hspace{4pt}}c} @{}}
\toprule
\textbf{Method} & \textbf{Backbone} & \textbf{Clip} & \textbf{Info} & \textbf{Paint} & \textbf{Quick} & \textbf{Real} & \textbf{Sketch} & \textbf{Avg}\\
\midrule
ERM        & \multirow{8}{*}{CNNs} & 58.1 (0.3) & 18.8 (0.3) & 46.7 (0.3) & 12.2 (0.4) & 59.6 (0.1) & 49.8 (0.4) & 40.9 \\
Mixup \cite{mixup}      &                        & 55.7 (0.3) & 18.5 (0.5) & 44.3 (0.5) & 12.5 (0.4) & 55.8 (0.3) & 48.2 (0.5) & 39.2 \\
CORAL \cite{coral}      &                        & 59.2 (0.1) & 19.7 (0.2) & 46.6 (0.3) & 13.4 (0.4) & 59.8 (0.2) & 50.1 (0.6) & 41.5 \\
MTL \cite{mtl}       &                        & 57.9 (0.5) & 18.5 (0.4) & 46.0 (0.1) & 12.5 (0.1) & 59.5 (0.3) & 49.2 (0.1) & 40.6 \\
SegNet \cite{segnet}    &                        & 57.7 (0.3) & 19.0 (0.2) & 45.3 (0.3) & 12.7 (0.5) & 58.1 (0.5) & 48.8 (0.2) & 40.3 \\
ARM \cite{arm}       &                        & 49.7 (0.3) & 16.3 (0.5) & 40.9 (1.1) &  9.4 (0.1) & 53.4 (0.4) & 43.5 (0.4) & 35.5 \\
Meta-DMoE \cite{metadmoe}  &                        & 63.5 (0.2) & 21.4 (0.3) & 51.3 (0.4) & 14.3 (0.3) & 62.3 (1.0) & 52.4 (0.2) & 44.2 \\
MABN \cite{wu2024mabn}       &                        & \textbf{64.2} & \textbf{23.6} & \textbf{51.5} & \textbf{15.2} & \textbf{64.6} & \textbf{54.1} & \textbf{45.5} \\
\midrule
DoPrompt \cite{doprompt}  & ViT-B/16 IMN           & 67.6 (0.2) & 24.6 (0.1) & 54.9 (0.1) & 17.5 (0.2) & 69.6 (0.3) & 55.2 (0.5) & 48.3 \\
\midrule
Zero-shot \cite{clip} & \multirow{6}{*}{\shortstack{ViT-B/16\\CLIP}} & 69.9 & 48.2 & 65.4 & 14.5 & 82.3 & 62.5 & 57.1 \\
ERM        &               & 68.0 (0.1) & 22.5 (0.6) & 46.5 (4.2) & 18.5 (0.9) & 58.7 (2.7) & 52.5 (1.2) & 44.4 \\
MIRO \cite{miro}       &                   & 74.9 (0.2) & 37.1 (0.4) & 59.8 (0.6) & \textbf{18.7 (1.2)} & 72.2 (0.2) & 61.2 (0.9) & 54.0 \\
VDPG \cite{vdpg}      &                    & 76.3 (0.2) & 49.3 (0.1) & 67.8 (0.1) & 17.4 (0.2) & 81.5 (0.3) & 66.6 (0.2) & 59.8 \\
L2C \cite{l2c} &           & 75.6 (0.1) & 52.1 (0.1) & 69.4 (0.1) & 17.3 (0.2) & 85.5 (0.1) & 67.1 (0.2) & 61.2 \\
\rowcolor{gray!10}
\textbf{DA-MergeLoRA} & & \textbf{76.66 (0.08)} & \textbf{54.49 (0.21)} & \textbf{70.91 (0.05)} & 17.87 (0.72) & \textbf{85.52 (0.12)} & \textbf{69.17 (0.08)} & \textbf{62.44 (0.16)} \\
\bottomrule
\end{tabular}
\end{adjustbox}
\caption{Evaluation of our method (DA-MergeLoRA) on the DomainNet dataset using the leave-one-out domain testing protocol.}
\label{tab:domainnet_results}
\end{subtable}

% ---------- (a) WILDS ----------
\begin{subtable}{\textwidth}
\centering
\footnotesize
\setlength{\tabcolsep}{6pt}
\renewcommand{\arraystretch}{1.0}
\begin{adjustbox}{max width=\textwidth}
\begin{tabular}{l l cc c cc cc}
\toprule

\multirow{2}{*}{\textbf{Method}} & 
\multirow{2}{*}{\textbf{Backbone}} &
\multicolumn{2}{c}{\textbf{iWildCam}} & 
\multicolumn{1}{c}{\textbf{Camelyon17}} &
\multicolumn{2}{c}{\textbf{FMoW}} &
\multicolumn{2}{c}{\textbf{PovertyMap}}\\

\cmidrule(lr){3-4} \cmidrule(lr){5-5} \cmidrule(lr){6-7} \cmidrule(lr){8-9}

& & Acc & Macro F1$^\dagger$ & Acc$^\dagger$ & WC Acc$^\dagger$ & Acc & WC Pearson r$^\dagger$ & Pearson r\\
\midrule

ERM        & \multirow{7}{*}{CNNs} & 71.6 (2.5) & 31.0 (1.3) & 70.3 (6.4) & 32.3 (1.25) & 53.0 (0.55) & 0.45 (0.06) & 0.78 (0.04)\\
CORAL \cite{coral}      &                       & 73.3 (4.3) & 32.8 (0.1) & 59.5 (7.7) & 31.7 (1.24) & 50.5 (0.36) & 0.44 (0.06) & 0.78 (0.05)\\
IRM \cite{irm}      &                       & 59.8 (3.7) & 15.1 (4.9) & 64.2 (8.1) & 30.0 (1.37) & 50.8 (0.13) & 0.43 (0.07) & 0.77 (0.05)\\
ARM-CML \cite{arm} &                       & 70.5 (0.6) & 28.6 (0.1) & 84.2 (1.4) & 27.2 (0.38) & 45.7 (0.28) & 0.37 (0.08) & 0.75 (0.04)\\
ARM-BN \cite{arm}     &                       & 70.3 (2.4) & 23.7 (2.7) & 87.2 (0.9) & 24.6 (0.04) & 42.0 (0.21) & 0.49 (0.21) & 0.84 (0.05)\\
Meta-DMoE \cite{metadmoe}  &                       & 77.2 (0.3) & 34.0 (0.6) & 91.4 (1.5) & 35.4 (0.58) & 52.5 (0.18) & 0.51 (0.04) & 0.80 (0.03)\\
MABN \cite{wu2024mabn}     &                       & \textbf{78.4 (0.6)} & \textbf{38.3 (1.2)} & \textbf{92.4 (1.9)} & \textbf{36.6 (0.41)} & \textbf{53.2 (0.52)} & 0.56 (0.05) & 0.84 (0.04)\\
\midrule
Zero-shot \cite{clip} & \multirow{4}{*}{\shortstack{ViT-B/16 \\ CLIP}} &  14.9 & 9.7 & 50.1 & 14.5 & 16.3 & 0.27 & 0.58\\
VDPG \cite{vdpg} &                 & 71.4 (0.2) & 30.1 (0.3) & 93.2 (0.3) & 37.8 (0.5) & 52.7 (0.3) & 0.38 (0.02) & 0.77 (0.02)\\
L2C \cite{l2c} &                  & 73.4 (0.4) & \textbf{35.2 (0.3)} & 94.2 (0.2) & 40.9 (0.4) & 54.8 (0.1) & 0.50 (0.02) & 0.80 (0.03)\\
\rowcolor{gray!10}
\textbf{DA-MergeLoRA} &  & \textbf{73.70 (1.49)} & 33.62 (1.63) & \textbf{96.90 (0.14)} & \textbf{52.26 (0.16)} & \textbf{57.60 (0.14)} & \textbf{0.56 (0.01)} & \textbf{0.84 (0.01)}\\
\bottomrule
\end{tabular}
\end{adjustbox}
\caption{Evaluation of our method (DA-MergeLoRA) on the WILDS datasets (iWildCam, Camelyon17, FMoW, and PovertyMap) under out-of-distribution (OOD) test conditions.}
\label{tab:wilds_results}
\end{subtable}
%------------------------------------------------
\label{tab:main_results}
\vspace{-1em}
\end{table}

\subsection{Main Results}

Table~\ref{tab:domainnet_results} and Table~\ref{tab:wilds_results} present the performance of our hypernetwork on DomainNet and WILDS, respectively. 

\subsubsection{DomainNet.} Compared to the state-of-the-art (SOTA) ViT-B/16 FSTT-DA method L2C \cite{l2c}, our approach improves average accuracy on DomainNet by $1.24\%$. Our model outperforms all methods across all domains except “quickdraw,” where it performs slightly lower ($-0.83\%$) than MIRO \cite{miro}. Relative to the SOTA CNN-based architecture MABN \cite{wu2024mabn}, our method improves average accuracy on DomainNet by $16.94\%$. 

\subsubsection{WILDS.} Compared to the SOTA ViT-B/16 FSTT-DA method L2C \cite{l2c}, our approach improves average accuracy by $2.80\%$ on FMoW, $2.70\%$ on Camelyon17, and $0.24\%$ on iWildCam, along with an $11.36\%$ gain in worst-case accuracy on FMoW. For the PovertyMap regression task, our method increases WC Pearson correlation by $0.06$ and average Pearson correlation by $0.04$. Our model outperforms L2C on all metrics except the iWildCam F1-score, where it performs slightly worse ($-1.58\%$). Relative to the SOTA CNN-based architecture MABN \cite{wu2024mabn}, our method improves average accuracy by $4.40\%$ on FMoW, $4.50\%$ on Camelyon17, and boosts worst-case accuracy on FMoW by $15.66\%$, while underperforming only on iWildCam and tying MABN on the PovertyMap regression task.

\begin{figure}[t]
    \centering

    % ---------- Row 1 ----------
    \begin{subfigure}[b]{0.47\textwidth}
        \centering
        \includegraphics[width=\textwidth]{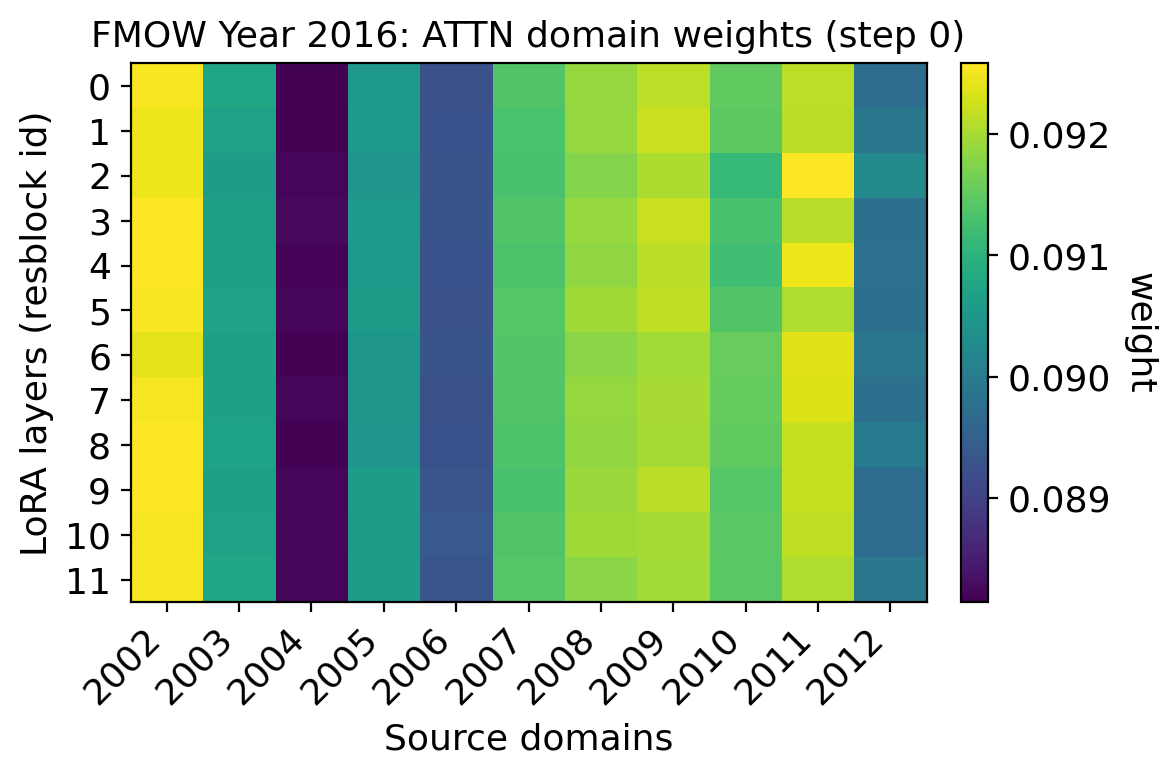}
        \caption{}
        \label{fig:a}
    \end{subfigure}
    \hfill
    \begin{subfigure}[b]{0.47\textwidth}
        \centering
        \includegraphics[width=\textwidth]{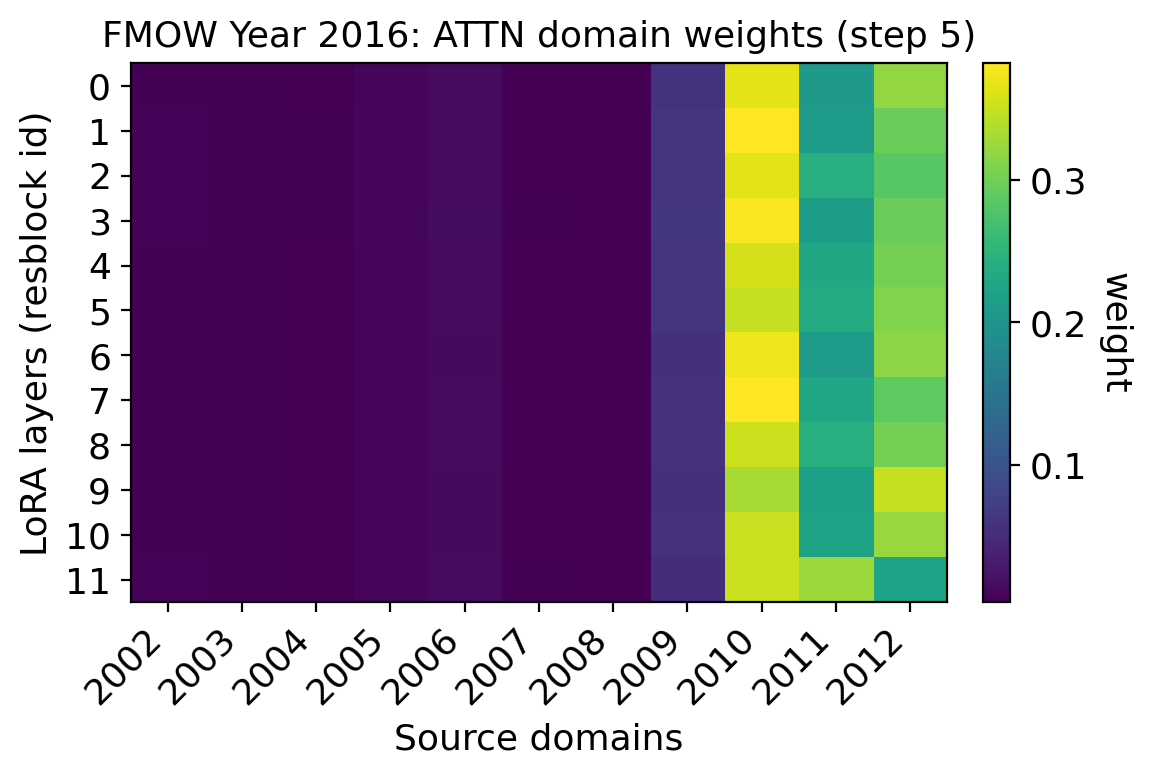}
        \caption{}
        \label{fig:b}
    \end{subfigure}

    % ---------- Row 2 ----------
    \begin{subfigure}[b]{0.47\textwidth}
        \centering
        \includegraphics[width=\textwidth]{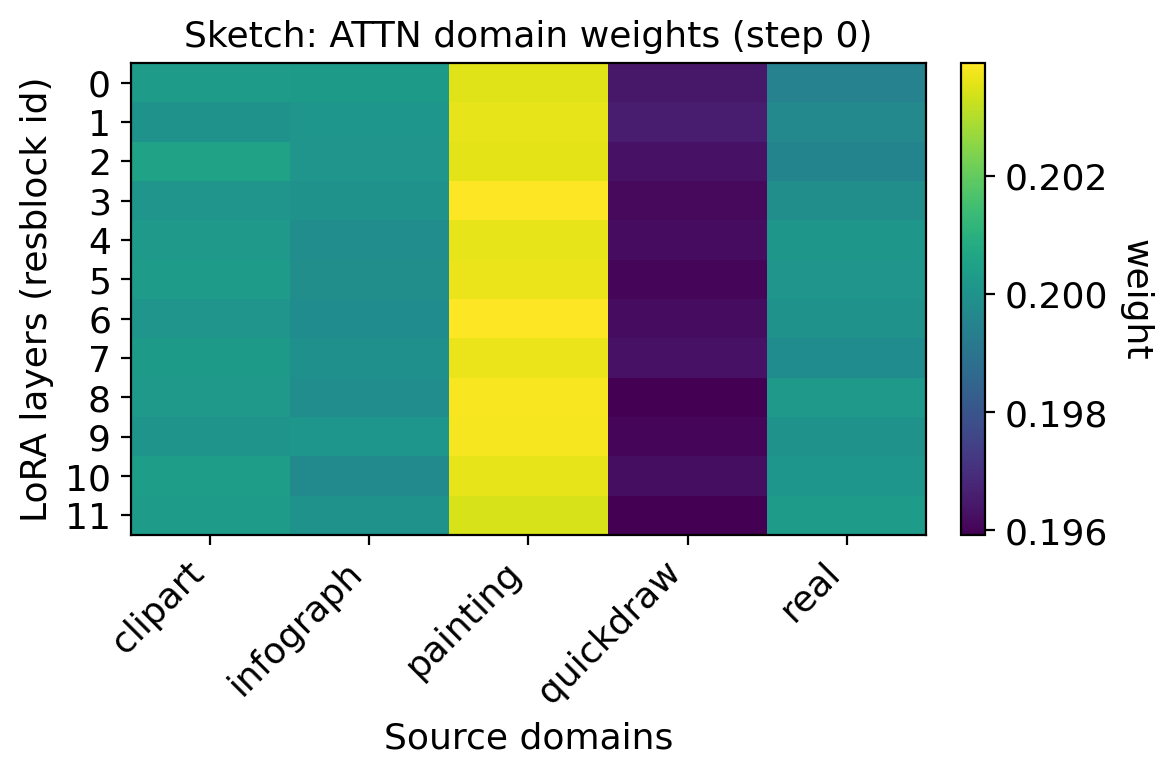}
        \caption{}
        \label{fig:c}
    \end{subfigure}
    \hfill
    \begin{subfigure}[b]{0.47\textwidth}
        \centering
        \includegraphics[width=\textwidth]{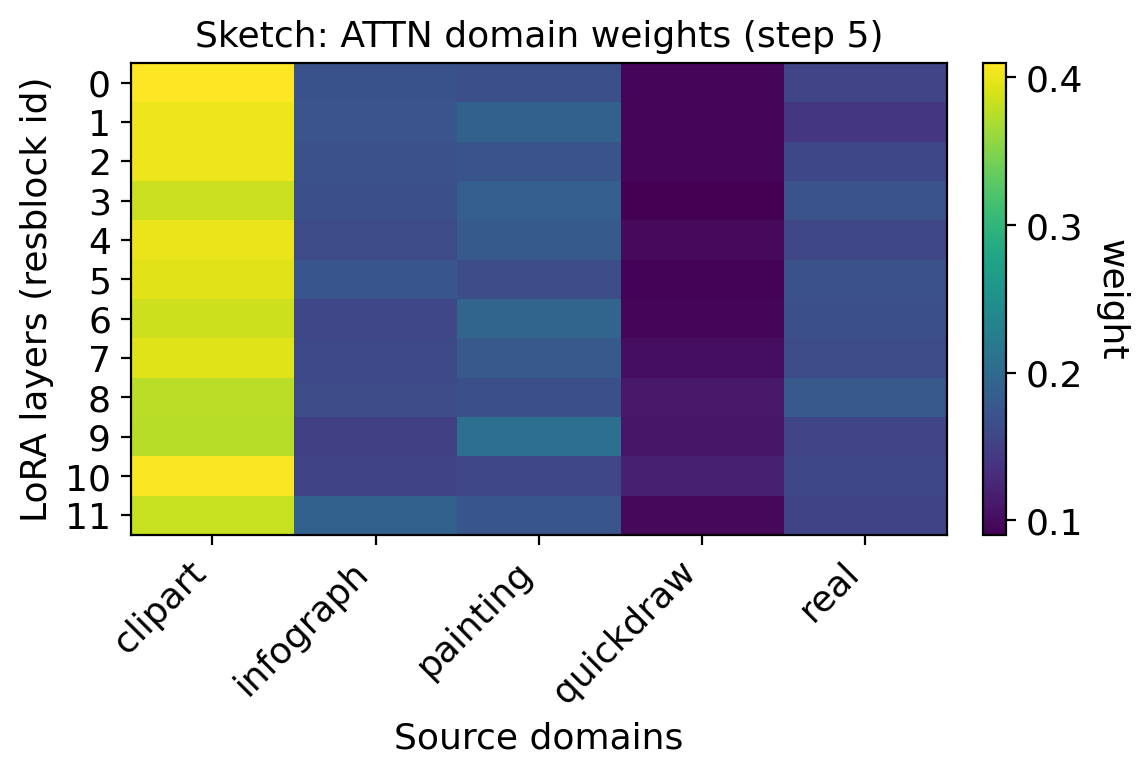}
        \caption{}
        \label{fig:d}
    \end{subfigure}

    \caption{\textbf{Heatmaps of average merge weights.} These plots illustrate the average merge weights per-attention-layer produced by the hypernetwork for each of the source domains. The x-axis shows the source domains and the y-axis shows the transformer layer. Plots (a) and (c) show the merge weights before training on FMoW ("Year 2016" target domain) and DomainNet ("Sketch" target domain). Plots (b) and (d) show results after training. Pre-training merge weights are nearly uniform; post-training, the hypernetwork prioritizes visually similar source domains over dissimilar or difficult domains.}
    \label{fig:merge_weights}
\end{figure}

\subsection{Ablation and Analyses}

We conduct analyses and ablation studies examining merge weights before and after hypernetwork training, evaluating various LoRA training techniques, and assessing the impact of different architectural components. Additional analyses are provided in the supplementary material (Appendix A.4).

\subsubsection{Merging weight heatmaps.} To visualize whether the hypernetwork learns meaningful merging weights, we plot heatmaps of average per-domain merging weights before and after training for 5 epochs. Fig. \ref{fig:merge_weights} shows results for FMoW ("Year 2016" target domain) and DomainNet ("Sketch" target domain). The x-axis denotes source domains (FMoW: years; DomainNet: image styles), and the y-axis denotes the CLIP transformer layers. Before training, weights are nearly uniform ($\sim0.09$ for FMoW, $\sim0.20$ for DomainNet). After training, the hypernetwork shifts merge weights towards domains visually similar to the target domain (later years for FMoW, "ClipArt" for "Sketch"), and down-weights dissimilar and difficult domains (earlier years for FMoW, "QuickDraw" for sketch), indicating it learns to emphasize relevant domains.

\begin{table}[t!]
\centering
\caption{Ablation experiments on the WILDS classification datasets. Results are reported as mean (standard deviation) over three trials with different random seeds. The $^\dagger$ symbol denotes the official dataset metric.}
\captionsetup[subtable]{position=top} 
% ---------- (a) WILDS ----------
\begin{subtable}{\textwidth}
\centering
\footnotesize
\setlength{\tabcolsep}{6pt}
\renewcommand{\arraystretch}{1.05}
\begin{adjustbox}{max width=\textwidth}
\begin{tabular}{l cc c cc}
\toprule
\textbf{Method} & \textbf{iWildCam Acc} & \textbf{iWildCam Macro F1$^\dagger$} & \textbf{Camelyon17 Acc$^\dagger$} & \textbf{FMoW WC Acc$^\dagger$} & \textbf{FMoW Acc} \\
\midrule
LoRA-universal   & 72.77 (3.13) & 28.88 (1.12) & 89.81 (6.89) & 51.02 (1.28) & 55.69 (0.24) \\
LoRA-average         & 14.98 (0.19) & 10.21 (0.05) & 95.26 (0.12) & 20.63 (0.08) & 21.95 (0.11) \\
LoRA-entropy-average & 41.89 (0.81) & 12.24 (0.22) & 95.28 (0.09) & 21.71 (0.50) & 23.15 (0.25) \\
LoRA-TIES & 49.85 (3.25) & 12.94 (0.84) & 95.95 (0.18) & 38.54 (1.27) & 43.98 (1.06)\\
LoRA-TIES-per-column & 51.25 (3.09) & 13.44 (0.20) & 95.35 (0.45) & 40.34 (1.12) & 45.35 (0.89)\\
LoRA-AdaMerge & 58.02 (0.92) & 18.66 (0.54) & 96.10 (0.63) & 44.85 (0.19) & 50.19 (0.08) \\
\rowcolor{gray!10}
\textbf{DA-MergeLoRA}    & \textbf{73.70 (1.49)} & \textbf{33.62 (1.63)} & \textbf{96.90 (0.14)} & \textbf{52.26 (0.16)} & \textbf{57.60 (0.14)}\\
\bottomrule
\end{tabular}
\end{adjustbox}
\caption{Comparison of LoRA training strategies: LoRA-universal (a single LoRA model trained on all source data combined), LoRA-average (simple parameter averaging), LoRA-entropy-average (few-shot entropy-weighted averaging), LoRA-TIES (TIES merging applied to all LoRA parameters at once), LoRA-TIES-per-column (TIES merging applied to each LoRA $\Delta W$ column separately), LoRA-AdaMerge (few-shot AdaMerging applied to LoRA parameters), and ours (hypernetwork-based merging).}
\label{tab:lora_train}
\end{subtable}
%-----------------------------------------------------------
%----------------------------------------------------------------
\begin{subtable}{\textwidth}
\footnotesize
\setlength{\tabcolsep}{6pt}
\renewcommand{\arraystretch}{1.1}
\centering
\begin{adjustbox}{max width=\textwidth}
\begin{tabular}{lccccc}
\toprule
\textbf{Method} & \textbf{iWildCam Acc} & \textbf{iWildCam Macro F1$^\dagger$} & \textbf{Camelyon17 Acc$^\dagger$}  & \textbf{FMoW WC Acc$^\dagger$} & \textbf{FMoW Acc} \\
\midrule
No layer and layer type token & 72.31 (0.43) & 28.95 (0.78) & 96.86 (0.34) &  49.16 (0.34) & 54.79 (0.05)\\
No domain token & 67.21 (1.50) & 25.72 (1.81) & \textbf{96.92 (0.12)} & 48.80 (0.38) & 54.16 (0.02) \\

No target images (noise) & 55.86 (1.00) & 17.39 (0.66) & 95.86 (0.48) &  43.04 (0.28)  & 48.43 (0.18)  \\
No target images (trainable vector)  & 63.47 (1.30) & 19.92 (0.65) & \textbf{96.92 (0.14)} & 51.98 (0.18) & 57.36 (0.13)\\
\rowcolor{gray!10}
\textbf{DA-MergeLoRA}  & \textbf{73.70 (1.49)} & \textbf{33.62 (1.63)} & 96.90 (0.14) & \textbf{52.26 (0.16)} & \textbf{57.60 (0.14)}\\
\bottomrule
\end{tabular}
\end{adjustbox}
\caption{Ablation studying the effect of:
(i) removing the layer and layer-type tokens, which indicate the transformer layer from which each LoRA column originates;
(ii) removing the domain tokens, which identify the source domain of each column;
(iii) providing random noise as input to the hypernetwork instead of target images for the support set; and
(iv) replacing the target-embedding branch with a trainable query vector that learns a general merging policy.
\label{tab:ablation}}
\end{subtable}
%-----------------------------------------------------------------------
%---------------------------------------------------------
\label{tab:ablation_experiments}
\end{table}

\subsubsection{LoRA training techniques.} We compare our method against five additional techniques for training LoRA on the source data: (i) LoRA-universal -- Train a single LoRA model on all source domains combined into one dataset. This yields a domain-invariant LoRA module, rather than domain-specific modules; (ii) LoRA-average -- Combine domain-specific LoRA modules by simple parameter-wise averaging; (iii) LoRA-entropy-average -- A simple few-shot method where we use a batch of target images to compute the entropy using each source LoRA model. The inverse entropy (normalized) is used as the merging weight, where lower entropy implies higher weight; (iv) LoRA-TIES and LoRA-TIES-per-column -- We implement the TIES \cite{ties} merging procedure to merge LoRA modules. We evaluate two variants: one that merges all LoRA $\Delta W$ matrices jointly, and one that merges individual $\Delta W$ columns separately (similar to our approach). We use the same hyperparameter settings recommended in the original paper: trim step K=20$\%$, sign election using resolve = mass, and disjoint merging using merge = dis-mean; (v) LoRA-AdaMerge -- An adaptation of AdaMerging \cite{adamerge} to our few-shot setting, where entropy minimization is performed to determine the merge weights for each LoRA model. 

Results are shown in Table~\ref{tab:lora_train} on the WILDS classification datasets. Our method outperforms the other LoRA techniques. On harder datasets (iWildCam, FMoW), simple averaging, entropy averaging, TIES, and AdaMerging perform markedly worse than the hypernetwork (between $15\%$ to $23\%$ decrease on iWildCam F1-score, and between $7\%$ to $32\%$ decrease on FMoW worst-case accuracy). On datasets with fewer source domains and more balanced per-domain classes (Camelyon17), simpler merging methods degrade less ($\sim2\%$ decrease), suggesting the hypernetwork excels in more complex merging situations. Our model also outperforms the universal LoRA in all cases, highlighting the benefit of domain-specific knowledge. 

\subsubsection{Component analysis.} 
We remove specific components of the hypernetwork to evaluate their impact on performance. The components examined are: (i) Layer and layer-type tokens -- We remove the learnable transformer layer ID token embeddings and layer-type (attention or projection) token embeddings appended to the LoRA columns; (ii) Domain tokens -- We remove the learnable domain-specific token embeddings appended to the LoRA columns; (iii) Conditional target images (replaced with noise) -- We remove the pseudo-target and target images as inputs to the hypernetwork during training and inference, replacing them with random noise; and (iv) Conditional target images (replaced with a trainable vector) -- We remove the target-image embedding branch and replace it with a trainable vector; instead of using pseudo-target images as the support set during training, we train a learnable query vector, which represents learning a general merging policy. 

Table~\ref{tab:ablation} reports the results on WILDS classification datasets. Removing the conditional target images reduces F1 performance by $13.7\%$–$16.2\%$ on iWildCam, which is the most difficult dataset with 48 target domains exhibiting significant domain shifts. Removing conditional images from FMoW also decreases performance, though not as severely. The learnable query vector reduces performance less than noise as input, indicating that a general merging policy can learn useful priors from the source domains. However, as seen in iWildCam, a domain-specific model performs better on complex domain shifts. Removing the layer tokens and domain tokens also lowers performance for both iWildCam and FMoW, showing that appending layer- and domain-specific tokens provides useful information. For Camelyon17, architectural modifications have negligible impact. As Camelyon17 is a less challenging dataset with performance nearing saturation ($\sim$96\%), the optimal merge weights are inherently near-uniform, as evidenced by the strong performance of simple parameter-wise averaging (Table~\ref{tab:lora_train}).

\subsubsection{Additional ablations and analyses.} 

We provide additional ablations and analyses in the supplementary material (Appendix A.4). These include architectural comparisons between MLP and cross-attention hypernetworks (Suppl. Table 2) and the impact of merging granularity (per-model, per-layer, and per-column) (Suppl. Table 3). Sensitivity analyses on support set batch size and number of source domains show our method remains stable across different scales (Suppl. Tables 4, 5). Analyzing the hardest cases for our hypernetwork (iWildCam F1-score, DomainNet "QuickDraw" accuracy), we find iWildCam's lower performance stems from class imbalance, while QuickDraw's stems from inherent domain difficulty. Qualitative examples and t-SNE visualizations further show improved class separation and error reduction over baseline CLIP (Suppl. Figs. 1, 2), with predicted merge weights positively correlating with domain similarity (Suppl. Fig. 3). Finally, computational analysis shows that DA-MergeLoRA is roughly 50\% faster and uses 50\% less memory at inference than prior FSTT-DA frameworks (Suppl. Table 7).

%-------------------------------------------------------------

\section{Conclusion}
In this work, we introduce a model-merging framework for few-shot test-time domain adaptation. We capture domain-specific information by training a LoRA module for each source domain. A hypernetwork, trained via meta-learning, then generates merging weights to optimally combine these modules for adaptation to a target domain, given a batch of unlabeled target images. Our method effectively adapts to novel domains and achieves state-of-the-art performance across diverse benchmarks. This work also opens several avenues for future research, including: (i) exploring merging strategies beyond meta-learning (e.g., reinforcement learning); (ii) developing mechanisms to handle class imbalance and novel classes; (iii) leveraging CLIP’s text encoder for stronger vision–language alignment; (iv) increasing the diversity of pseudo-target domains during meta-training through data augmentation or style mixing; (v) designing richer hypernetwork architectures; and (vi) employing LoRA column-alignment strategies to improve merging.

\section*{Acknowledgements}
This research was supported by the Natural Sciences and Engineering Research Council of Canada (NSERC).
\nocite{*}

\bibliographystyle{splncs04}
\bibliography{main}
\end{document}